%% file: main.tex
\documentclass[10pt,twocolumn,letterpaper]{article}

\usepackage[pagenumbers]{cvpr} 

\input{math_commands.tex}

\usepackage{graphicx}
\usepackage{amsmath}
\usepackage{amssymb}
\usepackage{booktabs}
\usepackage{float}
\usepackage{multirow}
\usepackage{colortbl}
\usepackage{xcolor}
\usepackage{color}
\usepackage[accsupp]{axessibility}

\usepackage[pagebackref,breaklinks,colorlinks]{hyperref}

\usepackage[capitalize]{cleveref}
\crefname{section}{Sec.}{Secs.}
\Crefname{section}{Section}{Sections}
\Crefname{table}{Table}{Tables}
\crefname{table}{Tab.}{Tabs.}

\makeatletter
\newcommand{\thickhline}{%
 \noalign {\ifnum 0=`}\fi \hrule height 1pt
 \futurelet \reserved@a \@xhline
}
\makeatother

\definecolor{aliceblue}{rgb}{0.94, 0.97, 1.0}

\def\confName{CVPR}
\def\confYear{2023}

\begin{document}

\title{ACSeg: Adaptive Conceptualization for Unsupervised Semantic Segmentation}

\author{
Kehan Li$^{1,3}$ \quad Zhennan Wang$^{2}$ \quad Zesen Cheng$^{1,3}$ \quad Runyi Yu$^{1,3}$ \quad Yian Zhao$^{5}$ \quad Guoli Song$^{2}$ \\
Chang Liu$^{4}$ \quad Li Yuan$^{1,2,3}$\thanks{Corresponding author. Project page: \href{https://lkhl.github.io/ACSeg}{https://lkhl.github.io/ACSeg}.} \quad Jie Chen$^{1,2,3}$$^*$ \and
\small{$^{1}$ School of Electronic and Computer Engineering, Peking University, Shenzhen, China} \quad
\small{$^{2}$ Peng Cheng Laboratory, Shenzhen, China} \\
\small{$^{3}$ AI for Science (AI4S)-Preferred Program, Peking University Shenzhen Graduate School, Shenzhen, China} \\
\small{$^{4}$ Department of Automation and BNRist, Tsinghua University, Beijing, China} \quad
\small{$^{5}$ Dalian University of Technology}
}

\maketitle

\begin{abstract}
Recently, self-supervised large-scale visual pre-training models have shown great promise in representing pixel-level semantic relationships, significantly promoting the development of unsupervised dense prediction tasks, e.g., unsupervised semantic segmentation (USS).
The extracted relationship among pixel-level representations typically contains rich class-aware information that semantically identical pixel embeddings in the representation space gather together to form sophisticated concepts. 
However, leveraging the learned models to ascertain semantically consistent pixel groups or regions in the image is non-trivial since over/ under-clustering overwhelms the conceptualization procedure under various semantic distributions of different images.
In this work, we investigate the pixel-level semantic aggregation in self-supervised ViT pre-trained models as image \textbf{Seg}mentation and propose the \textbf{A}daptive \textbf{C}onceptualization approach for USS, termed \textbf{ACSeg}.
Concretely, we explicitly encode concepts into learnable prototypes and design the \textbf{Adaptive} \textbf{Concept} \textbf{Generator} (ACG), which adaptively maps these prototypes to informative concepts for each image.
Meanwhile, considering the scene complexity of different images, we propose the modularity loss to optimize ACG independent of the concept number based on estimating the intensity of pixel pairs belonging to the same concept.
Finally, we turn the USS task into classifying the discovered concepts in an unsupervised manner.
Extensive experiments with state-of-the-art results demonstrate the effectiveness of the proposed ACSeg.
\end{abstract}

\section{Introduction}
\label{sec:intro}

\begin{figure}[t]
    \centering
    \includegraphics[width=\linewidth]{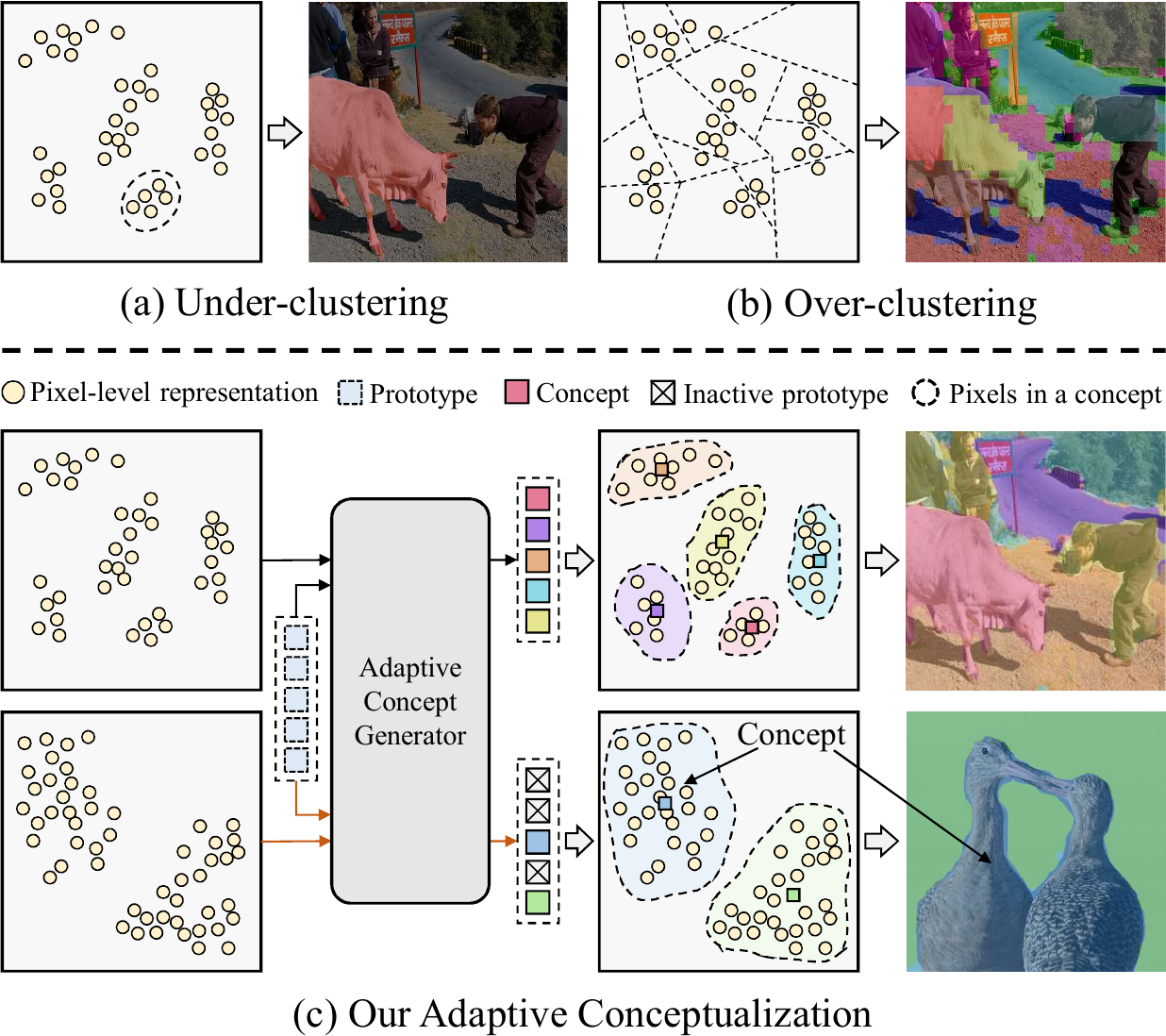}
    \caption{Comparison between existing methods and our adaptive conceptualization on finding underlying ``concepts" in the pixel-level representations produced by a pre-trained model. While under-clustering just focuses on a single object and over-clustering splits objects, our adaptive conceptualization processes different images adaptively through updating the initialized prototypes with the representations for each image.}
    \label{fig:intro}
    \vspace{-10pt}
\end{figure}

Semantic segmentation is one of the primary tasks in computer vision, which has been widely used in many domains, such as autonomous driving~\cite{geiger2013vision,cordts2016cityscapes} and medical imaging~\cite{sirinukunwattana2017gland,kumar2017dataset,fan2020pranet}.
With the development of deep learning and the increasing amount of data~\cite{everingham2012pascal,lin2014microsoft,cordts2016cityscapes,zhou2017scene}, uplifting performance has been achieved on this task by optimizing deep neural networks with pixel-level annotations~\cite{long2015fully}.
However, large-scale pixel-level annotations are expensive and laborious to obtain.
Different kinds of weak supervision have been explored to achieve label efficiency~\cite{shen2022survey}, \emph{e.g.}, image-level~\cite{ahn2018learning,wang2020self}, scribble-level~\cite{lin2016scribblesup}, and box-level supervision~\cite{oh2021background}.
More than this, some methods also achieve semantic segmentation without relying on any labels~\cite{hwang2019segsort,ji2019invariant}, namely unsupervised semantic segmentation (USS).

Early approaches for USS are based on pixel-level self-supervised representation learning by introducing cross-view consistency~\cite{ji2019invariant,cho2021picie}, edge detection~\cite{hwang2019segsort,zhang2020self}, or saliency prior~\cite{van2021unsupervised}.
Recently, the self-supervised ViT~\cite{caron2021emerging} provides a new paradigm for USS due to its property of containing semantic information in pixel-level representations.
We make it more intuitive through Figure~\ref{fig:intro}, which shows that in the representation space of an image, the pixel-level representations produced by the self-supervised ViT contain underlying clusters.
When projecting these clusters into the image, they become semantically consistent groups of pixels or regions representing ``concepts".

In this work, we aim to achieve USS by accurately extracting and classifying these ``concepts" in the pixel representation space of each image.
Unlike the previous attempts which only consider foreground-background partition~\cite{simeoni2021localizing,wang2022self,van2022discovering} or divide each image into a fixed number of clusters~\cite{melas2022deep,huang2022segdiscover}, we argue that it is crucial to consider different images distinguishably due to the complexity of various scenarios (Figure~\ref{fig:intro}).
We thus propose the \textbf{A}daptive \textbf{C}onceptualization for unsupervised semantic \textbf{Seg}mentation (ACSeg), a framework that finds these underlying concepts adaptively for each image and achieves USS by classifying the discovered concepts in an unsupervised manner.

\begin{figure}[t]
    \centering
    \includegraphics[width=\linewidth]{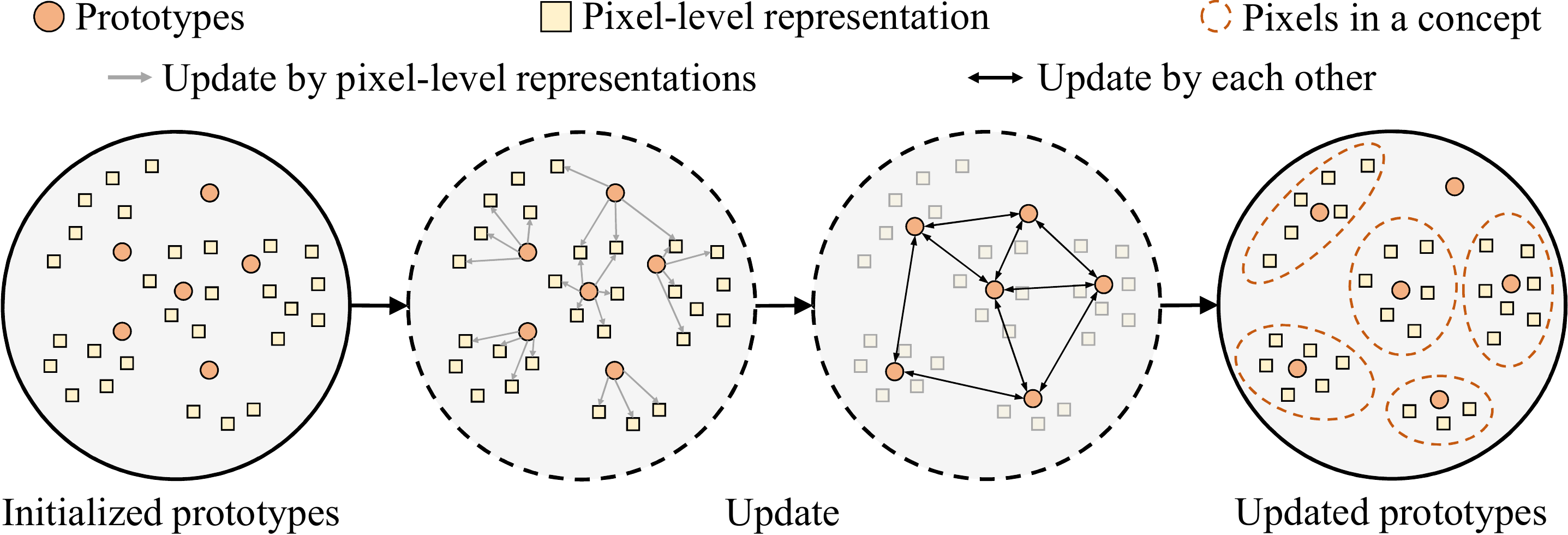}
    \caption{\textbf{Intuitive explanation for the basic idea of the ACG.} The concepts are explicitly encoded to learnable prototypes and dynamically updated according to the input pixel-level representations. After update, the pixels are assigned to the nearest concept in the representation space.}
    \label{fig:intuition}
    \vspace{-10pt}
\end{figure}

To achieve conceptualization, we explicitly encode concepts to learnable prototypes and adaptively update them for different images by a network, as shown in Figure~\ref{fig:intuition}.
This network, named as \textbf{A}daptive \textbf{C}oncept \textbf{G}enerator (ACG), is implemented by iteratively applying scaled dot-product attention~\cite{vaswani2017attention} on the prototypes and pixel-level representations in the image to be processed.
Through such a structure, the ACG learns to project the initial prototypes to the concept in the representation space depending on the input pixel-level representations.
Then the concepts are explicitly presented in the image as different regions by assigning each pixel to the nearest concept in the representation space.
The ACG is end-to-end optimized without any annotations by the proposed modularity loss.
Specifically, we construct an affinity graph on the pixel-level representations and use the connection relationship of two pixels in the affinity graph to adjust the strength of assigning two pixels to the same concept, motivated by the modularity~\cite{newman2004finding}.

As the main part of ACSeg, the ACG achieves precise conceptualization for different images due to its adaptiveness, which is reflected in two aspect:
Firstly, it can adaptively operate on pixel-level representations of different images thanks to the dynamic update structure.
Secondly, the training objective does not enforce the number of concepts, resulting in adaptive number of concepts for different images.
With these properties, we get accurate partition for images with different scene complexity via the concepts produced by the ACG, as shown in Figure~\ref{fig:intro}\textcolor{red}{(c)}.
Therefore, in ACSeg, the semantic segmentation of an image can finally be achieved by matting the corresponding regions in the image and classifying them with the help of powerful image-level pre-trained models.

For evaluation, we apply ACSeg on commonly used semantic segmentation datasets, including PASCAL VOC 2012~\cite{everingham2012pascal} and COCO-Stuff~\cite{lin2014microsoft,ji2019invariant}.
The experimental results show that the proposed ACSeg surpasses previous methods on different settings of unsupervised semantic segmentation tasks and achieves state-of-the-art performance on the PASCAL VOC 2012 unsupervised semantic segmentation benchmark without post-processing and re-training.
Moreover, the visualization of the pixel-level representations and the concepts shows that the ACG is applicable for decomposing images with various scene complexity.
Since the ACG is fast to converge without learning new representations and the concept classifier is employed in a zero-shot manner, we draw the proposed ACSeg as a generalizable method which is easy to modify and adapt to a wide range of unsupervised image understanding.

\section{Related Works}

\noindent \textbf{Vison Transformer.}
Transformer, a model mainly based on self-attention mechanism, is widely used in natural language processing~\cite{devlin2018bert,brown2020language} and cross-modal understanding~\cite{radford2021learning,li2022joint,li2022toward,jin2022expectationmaximization}.
Vison Transformer (ViT)~\cite{dosovitskiy2020image} is the first pure visual transformer model to process images.
Recently, Caron \emph{et al.}~\cite{caron2021emerging} propose self-\textbf{di}stillation with \textbf{no} labels (DINO) to train the ViT, and found a property that its features contain explicit information about the segmentation of an image.
Based on DINO, some previous studies~\cite{simeoni2021localizing,wang2022self,hamilton2021unsupervised,melas2022deep,van2022discovering,yin2022transfgu} successfully demonstrate extending this property to unsupervised dense prediction tasks.

\smallskip \noindent \textbf{Unsupervised Semantic Segmentation.}
With the development of self-supervised and unsupervised learning, unsupervised methods for semantic segmentation task start to emerge.
Among them, some methods focus on pixel-level self-supervised representation learning by introducing cross-view consistency~\cite{ji2019invariant,cho2021picie,zhang2021looking,ziegler2022self,ke2022unsupervised,wen2022self,wang2022fully}, visual prior~\cite{hwang2019segsort,zhang2020self,van2021unsupervised}, and continuity of video frames~\cite{bielski2022move}.
In contrast, Zadaianchuk \emph{et al.}~\cite{zadaianchuk2022unsupervised} adopt pre-trained object-centric representations and cluster them to segment objects.
Other methods exploit pixel-level knowledge of pre-trained generative models~\cite{melas2021finding} or self-supervised pre-trained convolutional neural network~\cite{wang2022freesolo,huang2022segdiscover}.
Recently, self-supervised ViTs trained with DINO has recently been explored for unsupervised dense prediction tasks due to the ability of representing pixel-level semantic relationships.
For semantic segmentation, Hamilton \emph{et al.}~\cite{hamilton2021unsupervised} train a segmentation head by distilling the feature correspondences, which further encourages pixel features to form compact clusters and learn better pixel-level representations.
TransFGU~\cite{yin2022transfgu} obtains semantic segmentation in a top-down manner by extracting class activate maps from DINO models.
Some approaches use the representations from DINO to segment images into regions.
Melas \emph{et al.}~\cite{melas2022deep} adopt spectral decomposition on the affinity graph to discover meaningful parts in an image and implement semantic segmentation of an image.
MaskDistill~\cite{van2022discovering} uses some hand-made rules based on pixel-level representations to find the salient region in an image.
In contrast, we aim at better extracting underlying concepts among the representations from DINO in an image by tackling the over/under-clustering problem.

\smallskip \noindent \textbf{Semantic Segmentation with Text.}
Vision-language pre-training models enable learning without annotations or zero-shot transfer on vision task~\cite{radford2021learning}.
For semantic segmentation, MaskCLIP~\cite{zhou2022maskclip} modifies the visual encoder of CLIP~\cite{radford2021learning} and applies the text-based classifier on pixel level.
Xu \emph{et al.}~\cite{xu2022groupvit} propose GroupViT, a hierarchical grouping vision transformer, and train it with image-to-text contrastive loss.
Finally, the semantic segmentation results can be obtained by the grouping result and text embeddings.
ReCo~\cite{shin2022reco} leverages the retrieval abilities of CLIP and the robust correspondences offered by modern image representations to co-segment entities.
Shin \emph{et al.}~\cite{shin2022namedmask} use CLIP to construct category-specific images and produce pseudo-label with a category-agnostic salient object detector bootstrapped from DINO.
In this paper, we also show that our method can be combined with recent vision-language pre-trained model to perform semantic segmentation with only text-image supervision.

\section{The Proposed ACSeg}

In this section, we describe the proposed method for USS in detail, starting from the whole framework.

\begin{figure*}[ht]
    \centering
    \includegraphics[width=\textwidth]{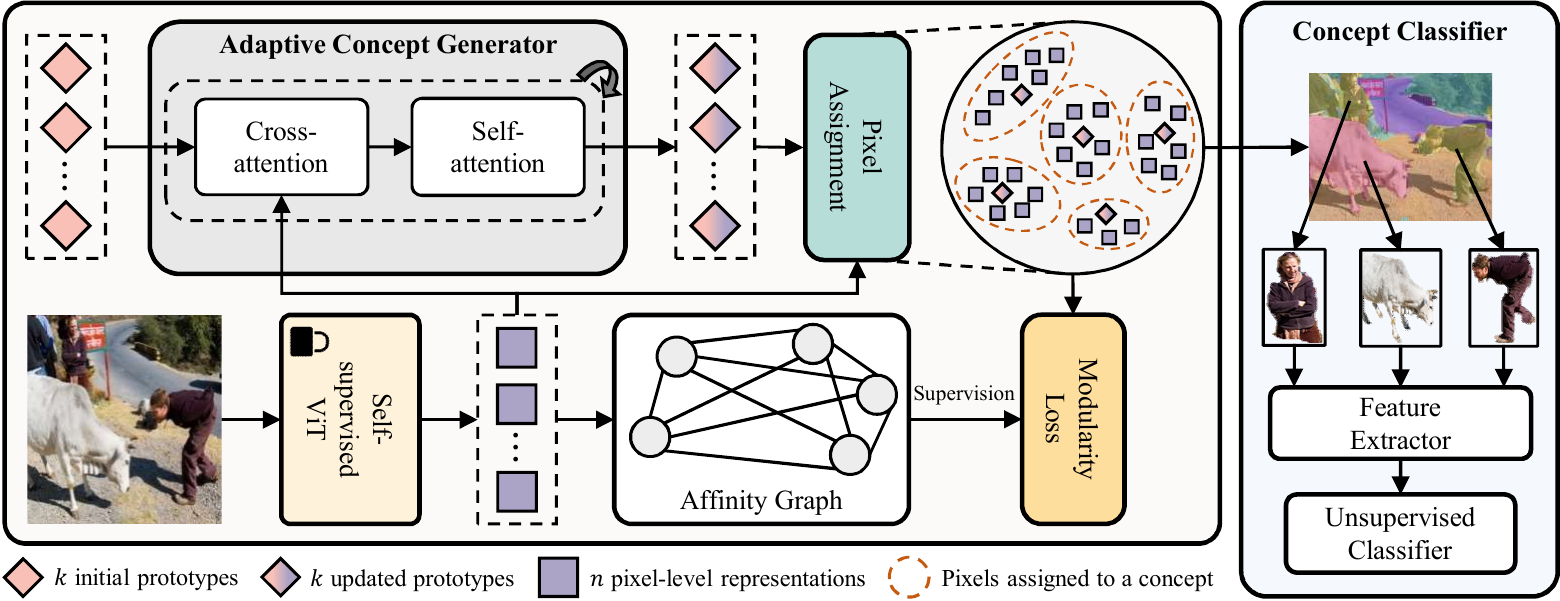}
    \caption{\textbf{Illustration of the proposed ACSeg.} For an image, we first use a self-supervised ViT to extract pixel-level representations, which imply the semantic relationship of pixels. The Adaptive Concept Generator (ACG) dynamically updates the initial prototypes to the underlying concepts in the representation space through scaled dot-product attention. Finally, the assignment of pixels is produced by the cosine similarity between pixel-level representations and the concepts, and the modularity loss is used to optimize the ACG. At last, the concept classifier is used to assign each concept to a pre-defined category thus obtain semantic segmentation of an image.}
    \label{fig:method}
    \vspace{-10pt}
\end{figure*}

\subsection{Overall Approach}

Figure~\ref{fig:method} illustrates the overall structure of ACSeg.
Starting with an image, we first apply a self-supervised ViT to generate pixel-level representations.
As mentioned above, these representations contain underlying concepts, which represents meaningful groups or regions of pixels.
The Adaptive Concept Generator (ACG) is designed to output the concepts explicitly.
Specifically, the ACG takes a series of learnable prototypes as input and iteratively updates them by interacting with the pixel-level representations, resulting in adaptive concept representations for each image.
Finally, the concepts are explicitly represented by pixel groups, which are obtained by assigning each pixel to the nearest concept in the representation space.

For optimization, we propose a novel loss function called modularity loss to train the ACG without any annotations.
Intuitively, the modularity loss works on pixel pairs.
We construct an affinity graph taking the pixel-level representations as vertices and their cosine similarity as edges.
The modularity loss calculates the intensity of two pixels belonging to the same concept using the metric defined in modularity~\cite{newman2004finding}, thus adjusting the concept representations.
At last, the concept classifier assigns each concept to a pre-defined category to obtain per-pixel class prediction, \emph{i.e.}, semantic segmentation of an image.
We introduce the details of each component in ACSeg in the following sections.

\subsection{Adaptive Concept Generator}

The role of ACG is to map the initial prototypes to the concept representations in each image.
Since the concept representations are different in different images and depend on the pixel-level representations of the image, we introduce the scaled dot-product attention~\cite{vaswani2017attention} to iteratively update the prototypes according to the pixel-level representations.
Specifically, we first apply cross-attention taking the prototypes as the \emph{query} and the pixel-level representations as the \emph{key} and \emph{value}.
Let $ \mC^l \in \sR^{k \times d} $ denote $ k $ prototypes after $ l $-th update and $ \mX \in \sR^{n \times d} $ denote $ n $ pixel-level representations from an image, the cross attention can be formulated as
\begin{equation}
    \bar{\mC^{l}} = \mathrm{Softmax}(\frac{\mC^{l-1} \mW_q (\mX \mW_k)^T}{\sqrt{d}})(\mX \mW_v),
\end{equation}
\begin{equation}
    \mC^{l} = \mC^{l-1} + \bar{\mC^{l}} \mW_o,
\end{equation}
where $ \mW_q, \mW_k, \mW_v, \mW_o \in \sR^{d \times d} $ are learnable linear projections.
The cross-attention updates prototypes adaptively with the pixel-level representations, which makes it possible to generate concepts adaptively for different images.

After that, self-attention is used to model the connections for different concepts.
Formulaically, it can be expressed as
\begin{equation}
    \bar{\mC^{l}} = \mathrm{Softmax}(\frac{\mC^{l-1} \mW_q (\mC^{l-1} \mW_k)^T}{\sqrt{d}})(\mC^{l-1} \mW_v),
\end{equation}
\begin{equation}
    \mC^{l} = \mC^{l-1} + \bar{\mC^{l}} \mW_o.
\end{equation}
The self-attention updates each prototype with other prototypes and makes it aware of the presence of others, for better adjusting their relative positions in the embedding space.

The ACG consists of $ N $ update steps, and each update step is made up of cross-attention, self-attention, and a Feed-forward Network~(FFN)~\cite{vaswani2017attention}.
With the attention mechanism, the ACG can learn the map from initial prototypes to concepts adaptively for different images.
For implementation, we adopt multi-head attention, layer normalization, and residual connection after the attention operation and the FFN, following the transformer~\cite{vaswani2017attention}.

\subsection{Pixel Assignment}
After ACG, each image has its own concepts.
Abstractly, each concept is a vector in the representation space, approximately the average of a group of gathered pixels.
Concretely, a concept consists of pixels with the same semantics.
This abstract-to-concrete transformation is achieved by assigning each pixel to a concept in ACSeg.

We first get a soft assignment for each pixel by calculating the cosine similarity with the concepts in the same image
\begin{equation}
    \emS_{i,j} = \cos<\vx_i,\vc_j>,
    \label{eq:soft_assignment}
\end{equation}
where $ \mS \in \sR^{n \times k}$ is the assignment matrix, $ \vx_i=\mX_{i,:} $ is the $ i $-th pixel embedding and $ \vc_j=\mC_{j,:} $ is the $ j $-th concept.
The soft assignment is differentiable and is used to optimize the network when training, which is described in Section~\ref{sec:loss}.
We assign each pixel to a definite concept during inference by the maximum similarity
\begin{equation}
    a_i = \underset{j}{\mathrm{argmax}}\, \cos<\vx_i,\vc_j>.
\end{equation}
By doing that, an image is segmented into $ m $ regions.
Each region is identified by a concept and consists of pixels assigned to this concept.
It is worth noting that $ m $ can be different for different images because the assignment is obtained by $ \mathrm{argmax} $ operation which do not guarantee that every concept is assigned at least once.
Due to the adaptive nature of this assignment and the adaptive generating of concepts for each image, we name the network Adaptive Concept Generator.

\subsection{Modularity Loss}
\label{sec:loss}

\begin{figure*}[!ht]
    \centering
    \includegraphics[width=\textwidth]{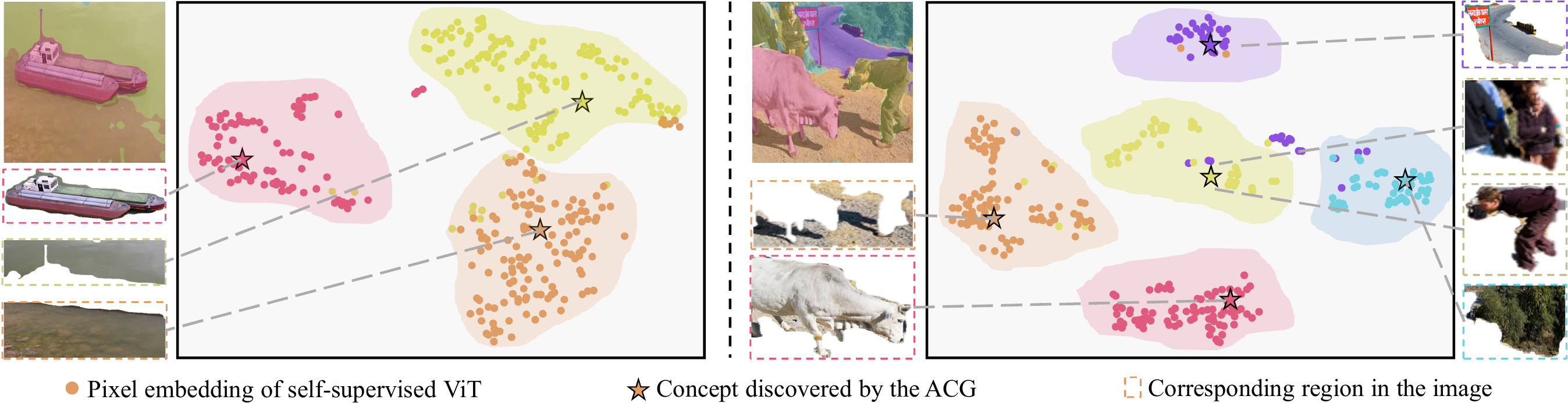}
    \caption{\textbf{t-SNE visualization of the pixel-level representations produced by self-supervised ViT and the corresponding concepts discovered by the ACG.} We mark the concepts found by the ACG in different colors. Based on the pixel-level representations, the ACG precisely finds the underlying concepts adaptively for different images with different scene complexity.}
    \label{fig:visualization}
    \vspace{-10pt}
\end{figure*}

For training the ACG, we design a loss function based on the idea of estimating the intensity of assigning two pixels to the same concept.
To achieve this goal, we introduce modularity~\cite{newman2004finding}, which is commonly used in community detection.
The modularity is built upon a graph, thus we first construct a fully connected undirected affinity graph for pixels from an image by taking them as vertices.
The weight of edge between two pixels which represents their affinity is calculated by the cosine similarity of them
\begin{equation}
    \emA_{i,j} = \max(0,\ \cos<\vx_i, \vx_j>).
\end{equation}
Here we truncate the value to a minimum of zero to avoid negative values in calculation.
Given two vertices $ i,j $, following the modularity, we estimate the intensity $ w_{ij} $ of assigning them to the same concept by
\begin{equation}
    w_{ij} = \emA_{i,j} - \frac{k_i \cdot k_j}{2m},
\end{equation}
where $ k_i = \sum_j \bm{A}_{i,j} $ is the sum of edges that are connected to vertex $ i $ and $ 2m = \sum_{i,j} \bm{A}_{i,j} $ is the sum of all edges in the graph.
For intuition, $ w_{ij} $ reflects the intensity of dividing pixels $ i $ and $ j $ into the same cluster via comparing the actual situation and the random situation.
When preserving the degrees of vertices in the graph but connecting vertices randomly, the probability of an edge existing between vertices $ i $ and $ j $ is $ k_i \cdot \frac{k_j}{2m} $.
Finally, $ w_{ij} $ reveals the possible existence of clusters by the comparison between the actual density of edges($ \emA_{ij} $) and the expected density when vertices are attached randomly ($ \frac{k_i \cdot k_j}{2m} $).
We refer readers to \cite{newman2004finding,fortunato2010community} for the detailed derivation.

For constructing a differentiable function, we now go back to the representation space and define the degree of two pixels belong to the same concept considering the upadated prototypes
\begin{equation}
    \delta(i,j) = \underset{c}{\max}\ \bar{\emS}_{i,c} \cdot \bar{\emS}_{j,c},\quad \bar{\mS} = \max(0, \mS),
    \label{eq:delta}
\end{equation}
where $ \mS $ is the soft assignment in~\eqref{eq:soft_assignment}, which makes the loss function differentiable.
We ignore values less than zero in $ S $ to avoid the case where both $ \emS_{i,c} $ and $ \emS_{j,c} $ are negative.
Since the pixel-level representations are fixed and only the prototypes are updated, the update of a prototype will be ambiguous if we take unrelated pixel pairs into account.
Therefore, we only choose one related prototype $ c $ for each pair by $ \underset{c}{\max}\ \bar{\emS}_{i,c} \cdot \bar{\emS}_{j,c} $ and calculating $ \delta(i,j) $ in the same way.
With $ w_{ij} $ and $ \delta(i,j) $, the modularity loss for pixels in an image can be formulated as
\begin{equation}
\begin{aligned}
    \mathcal{L} & = - \frac{1}{2m} \sum_{i,j} w_{ij} \delta(i,j) \\
    & = - \frac{1}{2m} \sum_{i,j}(\emA_{i,j} - \frac{k_i \cdot k_j}{2m}) \delta(i,j).
\end{aligned}
\end{equation}
Because we optimize the ACG with mini-batches, we use the sum of all edges $ 2m $ as the normalization factor for different images following the modularity.

Overall, the modularity loss adjusts the similarity of the prototypes to different pixel pairs according to the estimated intensity of $ i,j $ belonging to the same concept and then determines the position of updated prototypes in the embedding space.
Meanwhile, the $ \mathrm{max} $ function in~\eqref{eq:delta} adaptively chooses different cluster centers, which achieve the dynamic number of cluster centers for different images.
We summarize the reasons why the modularity loss is suitable for optimizing ACG as follows:
Firstly, no hyperparameters are required, making it robust to process images with scenarios of different complexity.
Secondly, the minimum of the modularity loss does not depend on the number of concepts, allowing the network to detect different counts of concepts in different images.
Through optimizing the modularity loss, the ACG finally learns to adaptively predict concepts in different images.

\begin{figure*}[!ht]
    \centering
    \includegraphics[width=\textwidth]{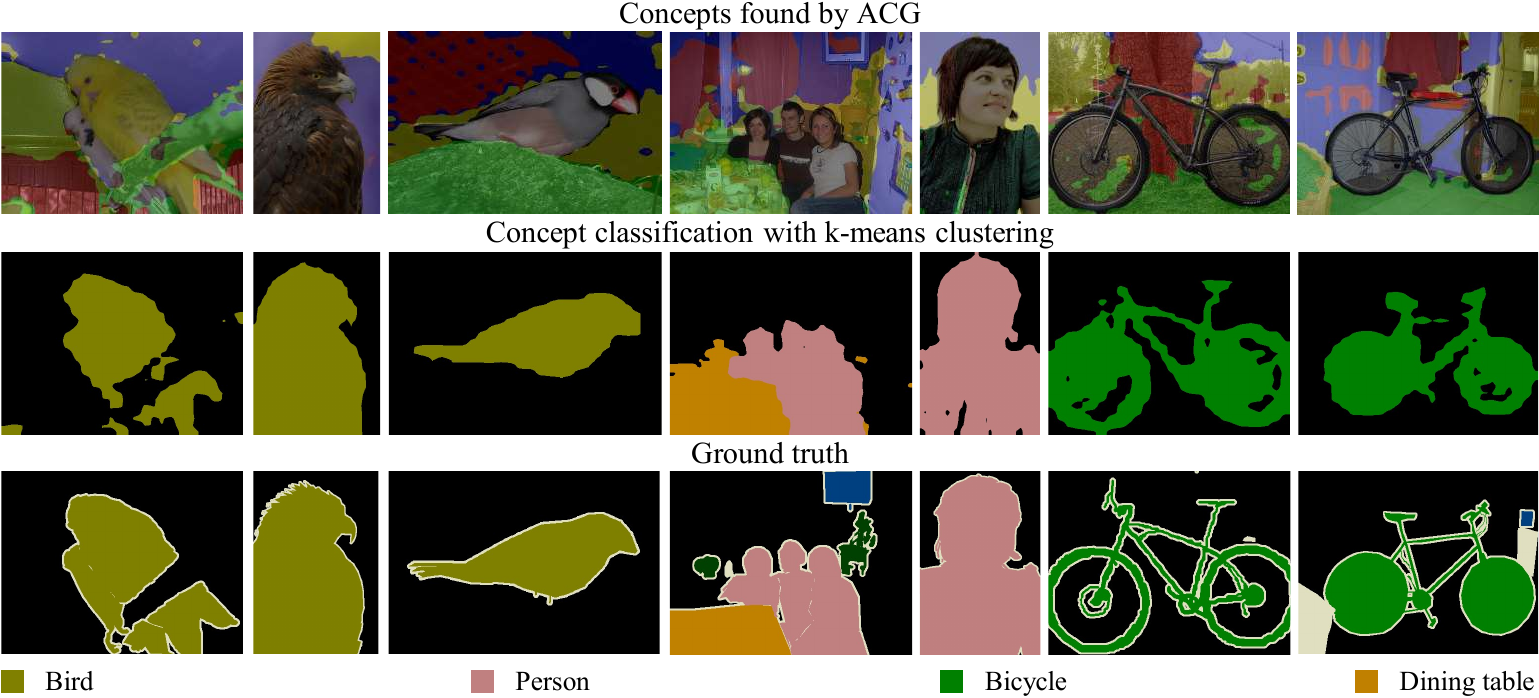}
    \caption{\textbf{Qualitative results on PASCAL VOC 2012 dataset.} We show the conceptualization results of ACG, semantic segmentation results based on k-means clustering and the ground truth in sequence. The proposed ACSeg can find different semantics within an image and realize precise semantic segmentation with image-level embedding from the self-supervised ViT.}
    \label{fig:qualitative_results}
    \vspace{-10pt}
\end{figure*}

\subsection{Concept Classifier}

In ACSeg, we finally obtain semantic segmentation of images by classifying the concepts produced by the ACG.
We first discuss the classifier for the special ``background" class, which is hard to describe when the labels are unavailable.
For instance, the background class actually contains a lot of sub-classes such as water, sky, land, wall, etc.
However, under the unsupervised setting, it is not known what or how many these sub-classes are.

We identify the background classes based on the attention of a self-supervised ViT, which is used when producing pixel-level representations so that no additional models are required.
Specifically, we first get a foreground score for each region by summing the attention values of pixels within it.
The attention values of pixels are from the last self-attention layer of the self-supervised ViT, taking the minimum of each attention head based on the idea that a pixel is likely to be foreground when it appears in at least one attention head.
With the foreground score of regions in an image, we cluster them into two categories and classify the cluster with a smaller score to background.

For the concepts belonging to foreground, the classification can be achieved by multiple ways.
One of them is to get a region-level representation for each concept by matting it from the original image and exploiting the ability to extract a discriminative image-level representation of the self-supervised ViT to get a region-level representation for each concept.
With these discriminative region-level representations, we apply the k-means clustering algorithm and $ k $-NN classifier on these representations to determine the class of each concept.
Besides, the recent progress of vision-language pre-trained models~\cite{radford2021learning} makes it possible to achieve unsupervised classification with the guidance of texts.
Therefore, we design a text-based classifier for the obtained concepts.
Specifically, we first obtain pixel-level representations using the visual encoder of the pre-trained model following MaskCLIP~\cite{zhou2022maskclip}.
Then the visual representation for each concept is produced by averaging the pixel-level representations within it.
Since the visual representations and language representations are aligned in CLIP~\cite{radford2021learning}, we use the text of the pre-defined categories as the classifier for each concept.

\section{Experiment}
\label{sec:experiment}

\subsection{Implementation Details}

We use ViT-Small~\cite{dosovitskiy2020image} trained with DINO~\cite{caron2021emerging} as the model to extract pixel-level and region-level representations.
From the ViT, we take tokens except the \emph{cls} token from the last layer as corresponding pixel-level representations.
For training and inference of the ACG, we resize the width and height of images to 224 and do not use additional data augmentation.
The ACG is optimized by AdamW~\cite{loshchilov2018decoupled} with a learning rate of 0.0001 and weight decay of 0.01.
We train the ACG for 2500 iterations using a batch size of 32.
We set the number of update steps in ACG to 6 and the number of prototypes to 5.
During inference, we use bilinear upsampling to restore the soft assignment of pixels to the original resolution before getting the hard assignment.
In the concept classifier, we first crop the image by the bounding box of the region and resize it to $ 224 \times 224 $.
After that, we mask out the pixels that are not part of the region after image normalization.
For qualitative evaluation, we adopt mean intersection over union (mIoU) and pixel accuracy as metrics, following most researches on semantic segmentation.
We conduct experiments on two commonly used semantic segmentation datasets PASCAL VOC 2012~\cite{everingham2012pascal} and COCO-Stuff~\cite{lin2014microsoft}, and we follow the previous works~\cite{ji2019invariant} to adopt the 27-class subset of COCO-Stuff.
Other detailed experiment settings are provided in Appendix.

\begin{figure*}[t]
    \centering
    \includegraphics[width=\textwidth]{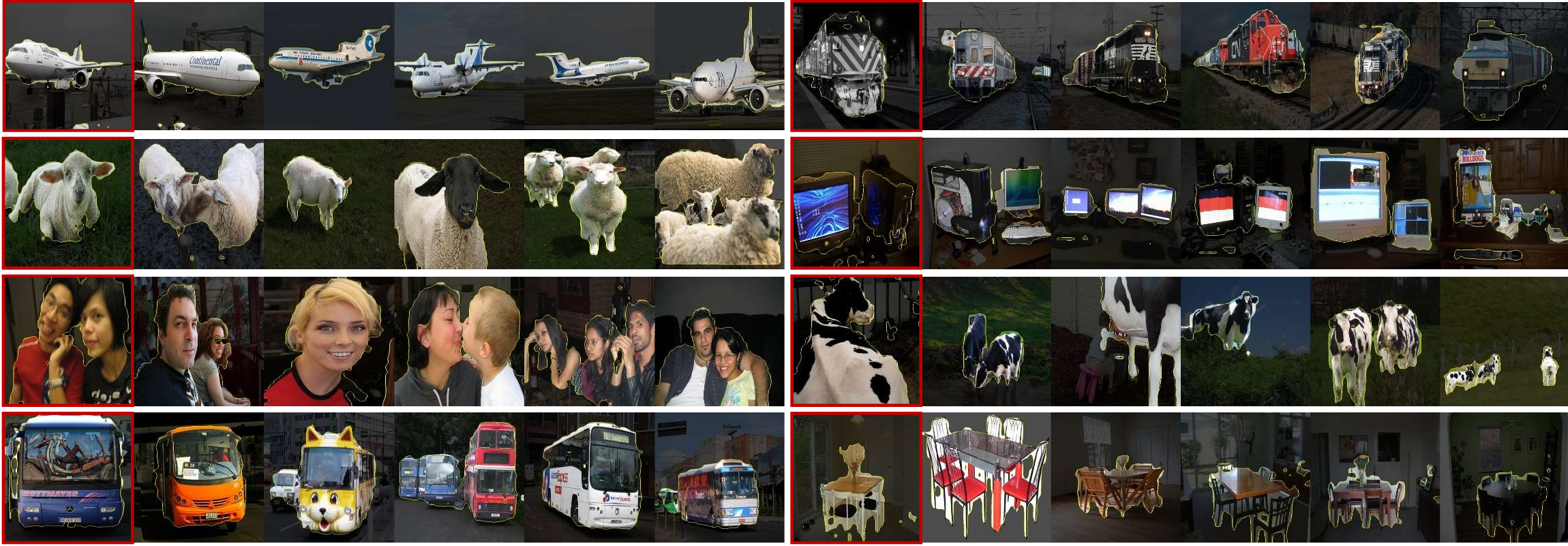}
    \caption{\textbf{Visualization of $ k $-NN retrieval results.} We show five concepts with the highest similarity following each query concept (with red frame). The concepts is shown by the highlighted area in the image.}
    \label{fig:knn}
\end{figure*}

\begin{table*}[ht]
    \centering
    \begin{minipage}{0.28\textwidth}
        \centering
        \renewcommand{\arraystretch}{1.1}
        \setlength{\tabcolsep}{5pt}
        \begin{tabular}{lc}
            \thickhline
            \bf Method & \bf mIoU \\
            \hline
            IIC \cite{ji2019invariant} & 9.8 \\
            MaskContrast \cite{van2021unsupervised} & 35.0 \\
            DSM\dag \cite{melas2022deep} & 37.2 $ \pm $ 3.8 \\
            Leopart \cite{ziegler2022self} & 41.7 \\
            TransFGU \cite{yin2022transfgu} & 37.2 \\
            MaskDistill \cite{van2022discovering} & 42.0 \\
            MaskDistill\dag \cite{van2022discovering} & 45.8 \\
            \hline
            \rowcolor{aliceblue!80} ACSeg (\emph{Ours}) & \textbf{47.1} $ \pm $ 2.4 \\
            \thickhline
        \end{tabular}
        \caption{\textbf{Unsupervised semantic segmentation results on PASCAL VOC.} \dag\ denotes results with re-training.}
        \label{tab:kmeans_voc}
    \end{minipage}
    \hfill
    \begin{minipage}{0.28\textwidth}
        \renewcommand{\arraystretch}{1.1}
        \setlength{\tabcolsep}{6.6pt}
        \begin{tabular}{lc}
            \thickhline
            \bf Method & \bf mIoU \\
            \hline
            MoCo v2 \cite{chen2020improved} & 4.4 \\
            IIC \cite{ji2019invariant} & 6.7 \\
            ImageNet \cite{he2016deep} & 8.9 \\
            DINO \cite{caron2021emerging} & 9.6 \\
            Modified DC \cite{cho2021picie} & 9.8 \\
            PiCIE \cite{cho2021picie} & 13.8 \\
            PiCIE+H \cite{cho2021picie} & 14.4 \\
            \hline
            \rowcolor{aliceblue!80} ACSeg (\emph{Ours}) & \textbf{16.4} $ \pm $ 0.9 \\
            \thickhline
        \end{tabular}
        \caption{\textbf{Unsupervised semantic segmentation results on COCO-Stuff-27 dataset.}}
        \label{tab:kmeans_coco}
    \end{minipage}
    \hfill
    \begin{minipage}{0.4\textwidth}
        \centering
        \renewcommand{\arraystretch}{1.1}
        \setlength{\tabcolsep}{6pt}
        \begin{tabular}{cccc}
            \thickhline
            \bf Dataset & \bf Method & \bf K=1 & \bf K=5 \\
            \hline
            \multirow{5}{*}{VOC} & MaskContrast \cite{van2021unsupervised} & 43.3 & - \\
            & DSM \cite{melas2022deep} & 32.1 & 31.9 \\
            & K-means & 45.1 & 49.1 \\
            & Spectral & 43.0 & 47.3 \\
            & \cellcolor{aliceblue!80}ACSeg (\emph{Ours}) & \cellcolor{aliceblue!80}\textbf{57.8} & \cellcolor{aliceblue!80}\textbf{61.0} \\
            \hline
            \multirow{3}{*}{COCO} & K-means & 29.9 & 33.1 \\
            & Spectral & 28.5 & 31.3 \\
            & \cellcolor{aliceblue!80}ACSeg (\emph{Ours}) & \cellcolor{aliceblue!80}\textbf{30.4} & \cellcolor{aliceblue!80}\textbf{34.0} \\
            \thickhline
        \end{tabular}
        \caption{\textbf{$k$-NN retrieval results on PASCAL VOC and COCO-Stuff-27.} Our ACSeg surpass the clustering baselines and previous methods.}
        \label{tab:knn}
    \end{minipage}
    \vspace{-11pt}
\end{table*}

\subsection{Qualitative Results}

To demonstrate the adaptiveness of the ACG, we first visualize the pixel-level representations and the concepts discovered by the ACG using t-SNE~\cite{van2008visualizing}, as shown in Figure~\ref{fig:visualization}.
Benefiting from the powerful representation capabilities of a ViT trained with DINO, the pixel-level representations reflect the semantic relationship between pixels, and pixels with the same semantics are aggregated into concepts.
It is obvious that the representations and counts of concepts are various for different images.
The ACG handles this situation correctly.
It accurately maps the initial prototypes to concepts in the pixel-level representation space for different images, which is shown by meaningful regions when projecting the assignment of each pixel to the original image.
Moreover, after optimizing by the proposed modularity loss, the ACG can produce an adaptive number of concepts for different images, which is achieved by dropping the concepts assigned to no pixels.

In addition, we show the segmentation results after the ACG and the concept classifier in the image in Figure~\ref{fig:qualitative_results}.
It can be found that the ACG achieves high-quality pixel-level localization (\emph{e.g.}, birds, fence, and tree branch in the first image) and can process scenes of varying complexity (\emph{e.g.}, there are more detected concepts in a complex scene (the first image) than the simple one (the second image), thanks to the flexibility of the proposed modularity loss).
The second row of Figure~\ref{fig:qualitative_results} shows the unsupervised semantic segmentation results by clustering the region-level representations.
Through clustering, concepts in different images are linked to further form region-level semantic groups.

\subsection{Quantitative Results}

\begin{table*}[ht]
    \centering
    \centering
    \begin{minipage}{0.33\textwidth}
        \centering
        \renewcommand{\arraystretch}{1.1}
        \setlength{\tabcolsep}{8pt}
        \begin{tabular}{lcc}
            \thickhline
            \multirow{2}{*}{\bf Method} & \multicolumn{2}{c}{\bf mIoU} \\
            \cline{2-3}
            & \bf VOC & \bf COCO \\
            \hline
            MaskCLIP \cite{zhou2022maskclip} & - & 19.6 \\
            GroupViT \cite{xu2022groupvit} & 51.2 & 20.3 \\
            ReCo \cite{shin2022reco} & - & 26.3 \\
            \hline
            \rowcolor{aliceblue!80}ACSeg (\emph{Ours}) & \textbf{53.9} & \textbf{28.1} \\
            \thickhline
        \end{tabular}
        \caption{\textbf{Comparison of unsupervised semantic segmentation with text.}}
        \label{tab:clip}
    \end{minipage}
    \hfill
    \begin{minipage}{0.34\textwidth}
        \centering
        \renewcommand{\arraystretch}{1.1}
        \setlength{\tabcolsep}{7pt}
        \begin{tabular}{lcc}
            \thickhline
            \bf Method & \bf mIoU & \bf Speed \\
            \hline
            K-Means \cite{hartigan1975clustering} & 28.6 & 2.4 \\
            Spectral \cite{von2007tutorial} & 28.3 & 3.4 \\
            AP \cite{dueck2009affinity} & 11.0 & 6.8 \\
            Agglomerative \cite{murtagh2012algorithms} & 13.9 & 15.8 \\
            \hline
            \rowcolor{aliceblue!80}ACG (\emph{Ours}) & \textbf{47.1} & \textbf{149.2} \\
            \thickhline
        \end{tabular}
        \caption{\textbf{Effectiveness and efficiency of the ACG.} Speed indicates images per second.}
        \label{tab:acg}
    \end{minipage}
    \hfill
    \begin{minipage}{0.28\textwidth}
        \centering
        \renewcommand{\arraystretch}{1.1}
        \setlength{\tabcolsep}{6pt}
        \begin{tabular}{ccc}
            \thickhline
            \bf Num & \bf Clustering & \bf Retrieval \\
            \hline
            2 & 35.6 & 43.7 \\
            7 & 46.9 & 56.9 \\
            10 & 42.1 & 54.4 \\
            15 & 34.5 & 51.1 \\
            \hline
            \rowcolor{aliceblue!80}5 & \textbf{47.1} & \textbf{57.8} \\
            \thickhline
        \end{tabular}
        \caption{\textbf{The effect of choosing different number of prototypes.}}
        \label{tab:num}
    \end{minipage}
    \vspace{-10pt}
\end{table*}

\noindent \textbf{K-means clustering.}
We first evaluate the performance of ACSeg for fully unsupervised semantic segmentation and show the results in Table~\ref{tab:kmeans_voc} and Table~\ref{tab:kmeans_coco}.
Following recent researches, we adopt k-means clustering on region-level representations to classify foreground regions and evaluate the quality of clusters with ground truth via Hungarian matching~\cite{kuhn1955hungarian}.
Since the k-means algorithm is greatly affected by the initial value of cluster centers, we run it ten times and report the results by $ mean \pm std $.
Comparing with the methods which also adopt the region clustering pipeline~\cite{melas2022deep,van2022discovering}, we reach higher segmentation performance without re-training and achieve state-of-the-art performance on PASCAL VOC 2012 dataset, demonstrating the superiority of the ACSeg brought by its adaptive conceptualization process.
In addition, the proposed ACSeg can be viewed as a transfer from self-supervised image-level models to dense prediction tasks, which only exploit the extracted representations of the pre-trained models rather than learning new representations.
Therefore, it only requires an extremely small cost and only takes tens of minutes to train and get segmentation results.
To this end, it has the advantage of being flexible and easy to use while achieving leading performance compared to the methods that train a segmentation models from scratch.

\noindent \textbf{$ k $-NN retrieval.}
We evaluate the effect of ACSeg method with $ k $-NN classifier, to show the quality of region-level representation and the localization quality of the concepts generated by the ACG.
In this setting, the class prediction of each concept in the validation set is obtained by the label of its nearest concepts in the training set, and the label of each concept in the training set is set to the same as the most overlapping ground-truth region.
For comparison, we report the results of previous works~\cite{van2021unsupervised,melas2022deep} which produce region-level representations and some clustering baselines (k-means clustering and spectral clustering) which replace the ACG in our method, as shown in Table~\ref{tab:knn}.
To intuitively show the retrieval performance, we also show some concepts and their five nearest neighbors in Figure~\ref{fig:knn}.
The proposed ACSeg surpasses the previous methods and baselines, showing the promising performance of fully unsupervised region-level representations.

\noindent \textbf{Unsupervised semantic segmentation with text.}
The emergence of vision-language pre-training models helps the unsupervised classification in visual tasks by constructing classifiers from prompt texts.
MaskCLIP~\cite{zhou2022maskclip} proposes to modify the visual encoder of CLIP~\cite{radford2021learning} and apply the text-based classifiers to pixel level.
However, the pixel-level classification results are relatively coarse due to the image-level pre-training task.
Here we demonstrate the effectiveness of combining the localization ability of the discovered concepts and the classification ability of CLIP through Table~\ref{tab:clip}.
With the help of accurate localization of concepts, we get higher performance than the related works MaskCLIP~\cite{zhou2022maskclip}, GroupViT~\cite{xu2022groupvit} and ReCo~\cite{shin2022reco}.

\subsection{Ablation Study}

\noindent \textbf{Effectiveness of the ACG.}
As the crucial part of ACSeg, we discuss the effectiveness of the ACG here on PASCAL VOC dataset.
Since the ACSeg can be regarded as a kind of clustering, we adopt some commonly used clustering methods k-means~\cite{hartigan1975clustering}, spectral clustering~\cite{von2007tutorial}, affinity propagation~\cite{dueck2009affinity}, and agglomerative clustering~\cite{murtagh2012algorithms}.
The results in Table~\ref{tab:acg} demonstrate the effectiveness of the ACG.
The ACG is more suitable for clustering pixel-level representations of produced by self-supervised ViT due to adaptively considering the attributes in different images, so that achieving better performance.
Thanks to the adaptiveness of the ACG, it can process all images through a single network.
Therefore, it is faster and can be highly accelerated by GPU compared to the classical iterative clustering algorithms.

\smallskip \noindent \textbf{The number of prototypes} is the main design choice in our method.
Although the ACG generates an adaptive number of concepts, the number of prototypes impacts on the granularity of conceptualization.
Specifically, it determines the upper bound on the concepts detected in each image, and more prototypes lead to finer-grained conceptualization for all images.
We show the clustering and retrieval results on VOC with different numbers of prototypes in Table~\ref{tab:num}.
When setting the number to 2, the ACG degenerates into foreground-background segmentation, resulting in performance degradation.
A large value also degrades the performance since over-clustering may occur on some objects with finer-grained division.
Our method is robust when the value is within a reasonable range.

\section{Conclusion and Discussion}
In this work, we propose the ACSeg, an approach that efficiently transfers the pixel-level knowledge of self-supervised ViT for unsupervised semantic segmentation.
We design the ACG to achieve adaptive conceptualization, \emph{i.e.}, generating concepts from pixel-level representations while considering the semantic distributions of different images.
Meanwhile, a novel loss function called modularity loss is designed to optimize the ACG without relying on any annotations and enable an adaptive number of concepts in different images.
Finally, we turn the USS task into classifying the discovered concepts in an unsupervised manner.
Qualitative and quantitative experimental results under different settings demonstrate the effectiveness and superiority of the proposed method as a directly transfer for self-supervised ViT without learning new representations.

\medskip \noindent \textbf{Acknowledgements.}
This work was supported in part by Natural Science Foundation of China (No. 61972217, 32071459, 62176249, 62006133, 62271465, 62202014), and the Natural Science Foundation of Guangdong Province in China (No. 2019B1515120049).
Li Yuan was supported in part by the National Key R\&D Program of China (2022ZD0118101) and also sponsored by CCF Tencent Open Research Fund.

{\small
\bibliographystyle{ieee_fullname}
\bibliography{reference}
}

\clearpage
\onecolumn
\appendix

\begin{center}
    \Large \bf Appendix
\end{center}

\section{Limitations}

The proposed ACSeg directly exploits the pixel-level representations of a pre-trained ViT model.
Although the ACG accurately groups pixels to concepts in the representation space, it can not be guaranteed that all representations reflect the corresponding semantic relationship unambiguously, especially when there is a gap between the pre-training dataset and the downstream task.
On the other hand, the region-level representation is also transferred from pre-trained models and thus suffers from the domain shift.
Although it performs well on VOC, the gap between the pre-training data of the backbone~(ImageNet) and COCO causes only modest performance on COCO.
It can be mitigated by training a task-specific model to produce better representations like STEGO~\cite{hamilton2021unsupervised} and SlotCon~\cite{wen2022self}.
We take solving this issue as future research.
Meanwhile, the pre-defined number of prototypes is neccesary as the optimal partition of images is unknowable without given granularity.
This hyperparameter impacts the granularity since each pixel pair is assigned to the closest prototype when optimizing the loss.
Empirically, our method performs well when the variance of image complexity is not so large and this hyperparameter is determined by observation on several samples.
Dealing with extremely large and complex datasets is still a limitation.

\section{Additional Implementation Details}

\noindent \textbf{Dataset.}
We use the PASCAL VOC 2012~\cite{everingham2012pascal} dataset (with extra augmentation data~\cite{hariharan2011semantic}) and COCO-Stuff~\cite{lin2014microsoft} dataset for training and evaluation.
For the COCO-Stuff dataset, we exploit the 27-classes subset and the ``curated" split\footnote{\url{https://www.robots.ox.ac.uk/~xuji/datasets/COCOStuff164kCurated.tar.gz}} introduced by IIC~\cite{ji2019invariant}.

\smallskip \noindent \textbf{Baseline.}
Since the ACSeg can be regarded as a kind of clustering, we adopt some commonly used clustering methods k-means~\cite{hartigan1975clustering}, spectral clustering~\cite{von2007tutorial}, affinity propagation~\cite{dueck2009affinity}, and agglomerative clustering~\cite{murtagh2012algorithms} as baselines for comparison.
We use the implementation of these algorithms in Scikit-learn~\cite{pedregosa2011scikit}.
For these baselines, it is difficult to choose a fixed set of parameters for all images, which is why these methods cannot achieve good adaptiveness.
We chose relatively suitable hyperparameters for different baselines, as shown in Table~\ref{tab:appendix_hyperparameters}.
On the other hand, their are some existing over/ under-clustering methods for replacing the ACG, such as LOST~\cite{simeoni2021localizing} and DSM~\cite{melas2022deep}.
We compare the proposed ACG with them and show the results in Table~\ref{tab:appendix_acg}.

\smallskip \noindent \textbf{K-means Clustering.}
In this setting, we run k-means clustering on the region-level representations produced by the concept classifier to get the class prediction of each concept.
For the VOC 2012 dataset, we first recognize the concepts belonging to background as mentioned in Section 3.5.
After that, we run k-means to assign the representations of predicted foreground concepts to 20 clusters and finally get predictions of 21 classes (20 foreground classes + 1 background class).
The background class is recognized by the method proposed in Section 3.5.
We show the results with some other possible alternatives in Table~\ref{tab:appendix_background}.
For the COCO-Stuff dataset, since there is no background category, we directly cluster all representations into 27 classes.
The evaluation is done by matching the predicted clusters with the ground truth by Hungarian algorithm~\cite{kuhn1955hungarian}.

\begin{table}[h]
    \centering
    \begin{minipage}{0.41\textwidth}
        \centering
        \small
        \renewcommand{\arraystretch}{1.2}
        \setlength{\tabcolsep}{11pt}
        \begin{tabular}{ccc}
            \thickhline
            \textbf{LOST \cite{simeoni2021localizing}} & \textbf{DSM \cite{melas2022deep}} & \cellcolor{aliceblue}\textbf{ACG (\emph{Ours})} \\
            \hline
            18.2 & 36.8 & \cellcolor{aliceblue}\textbf{47.1} \\
            \thickhline
        \end{tabular}
        \caption{\textbf{Results of other baselines.}}
        \label{tab:appendix_acg}
    \end{minipage}
    \hfill
    \begin{minipage}{0.58\textwidth}
        \centering
        \small
        \renewcommand{\arraystretch}{1.2}
        \setlength{\tabcolsep}{6.6pt}
        \begin{tabular}{ccc}
            \thickhline
            \textbf{Max Area \cite{melas2022deep}} & \textbf{Unsupervised Saliency \cite{wang2022self}} & \cellcolor{aliceblue}\textbf{Attention (\emph{Ours})} \\
            \hline
            39.1 & 46.0 & \cellcolor{aliceblue}\textbf{47.1} \\
            \thickhline
            \end{tabular}
            \caption{\textbf{Results of other background classification methods.}}
        \label{tab:appendix_background}
    \end{minipage}
\end{table}

\smallskip \noindent \textbf{$ k $-NN Retrieval.}
We adopt the weighted $ k $-NN classifier in this setting.
Specifically, the soft label of a concept is calculated by weighted averaging one-hot labels of $ k $ most similar concepts by their similarity, where we use the cosine distance between region embeddings as the similarity.
Finally, the category with the highest score in the soft label is used as the classification result of a concept.
We generate labels for concepts in the training set by the most overlapping ground truth region.
The evaluation is done on the \emph{val} set of VOC 2012 and COCO-Stuff.
For the VOC 2012 dataset, we chose the \emph{train} and \emph{aug}\footnote{samples in SBD~\cite{hariharan2011semantic} but not in \emph{train} and \emph{val} sets} sets as the training set.
For the COCO-Stuff dataset, we only report the results produced by using the first $ 10k $ samples of the \emph{train} set in the main text, because it is very time-consuming to get the results of baselines.
We show the results of our method when using all samples of the \emph{train} set in Table~\ref{tab:appendix_knn}.

\begin{table}[h]
    \centering
    \begin{minipage}{0.615\textwidth}
        \centering
        \setlength{\tabcolsep}{4pt}
        \begin{tabular}{ll}
            \thickhline
            \bf Algorithm & \bf Hyperparameters \\
            \hline
            K-means & n\_clusters = 5, init = `k-means++' \\
            Spectral clustering & n\_clusters = 5, n\_components = 5 \\
            Affinity propagation & damping=0.5, preference = -2\\
            Agglomerative clustering & distance\_threshold = 0.65, linkage = `average' \\
            \thickhline
        \end{tabular}
        \caption{\textbf{Hyperparameters for different clustering baselines k-means, spectral clustering, affinity propagation, and agglomerative clustering.}}
        \label{tab:appendix_hyperparameters}
    \end{minipage}
    \hfill
    \begin{minipage}{0.36\textwidth}
        \centering
        \setlength{\tabcolsep}{6pt}
        \begin{tabular}{cccc}
            \thickhline
            \bf Dataset & \bf Method & \bf K=1 & \bf K=5 \\
            \hline
            \multirow{4}{*}{COCO} & K-means & 29.9 & 33.1 \\
            & Spectral & 28.5 & 31.3 \\
            & \cellcolor{aliceblue!80}ACSeg (Ours) & \cellcolor{aliceblue!80}30.4 & \cellcolor{aliceblue!80}34.0 \\
            & \cellcolor{aliceblue!80}ACSeg\dag (Ours) & \cellcolor{aliceblue!80}\textbf{33.8} & \cellcolor{aliceblue!80}\textbf{37.7} \\
            \thickhline
        \end{tabular}
        \caption{\textbf{Additional $k$-NN retrieval results.} \dag\ indicates using all samples.}
        \label{tab:appendix_knn}
    \end{minipage}
\end{table}

\smallskip \noindent \textbf{Semantic Segmentation with Text.}
We first generate the text-based classifiers using the text encoder of CLIP~\cite{radford2021learning} and the pre-defined categories.
Following \cite{radford2021learning,zhou2022maskclip}, the words of categories are wrapped to sentences by templates and the classifier for a category is the average of the corresponding wrapped sentences.
For the VOC 2012 dataset, we use the background classifier mentioned in Section 3.5 and only construct the text-based classifier for 20 foreground categories.
For the COCO-Stuff dataset, we first classify concepts to all the things and stuff categories defined in COCO and then map them to 27 classes following ReCo~\cite{shin2022reco}.
For the visual representations, we first get the pixel-level representations following MaskCLIP~\cite{zhou2022maskclip} with CLIP-ViT-B/16 and then produce the region-level representation for each concept by averaging the pixels within it.

\section{Additional Qualitative Results}

\begin{figure*}[ht]
    \includegraphics[width=\textwidth]{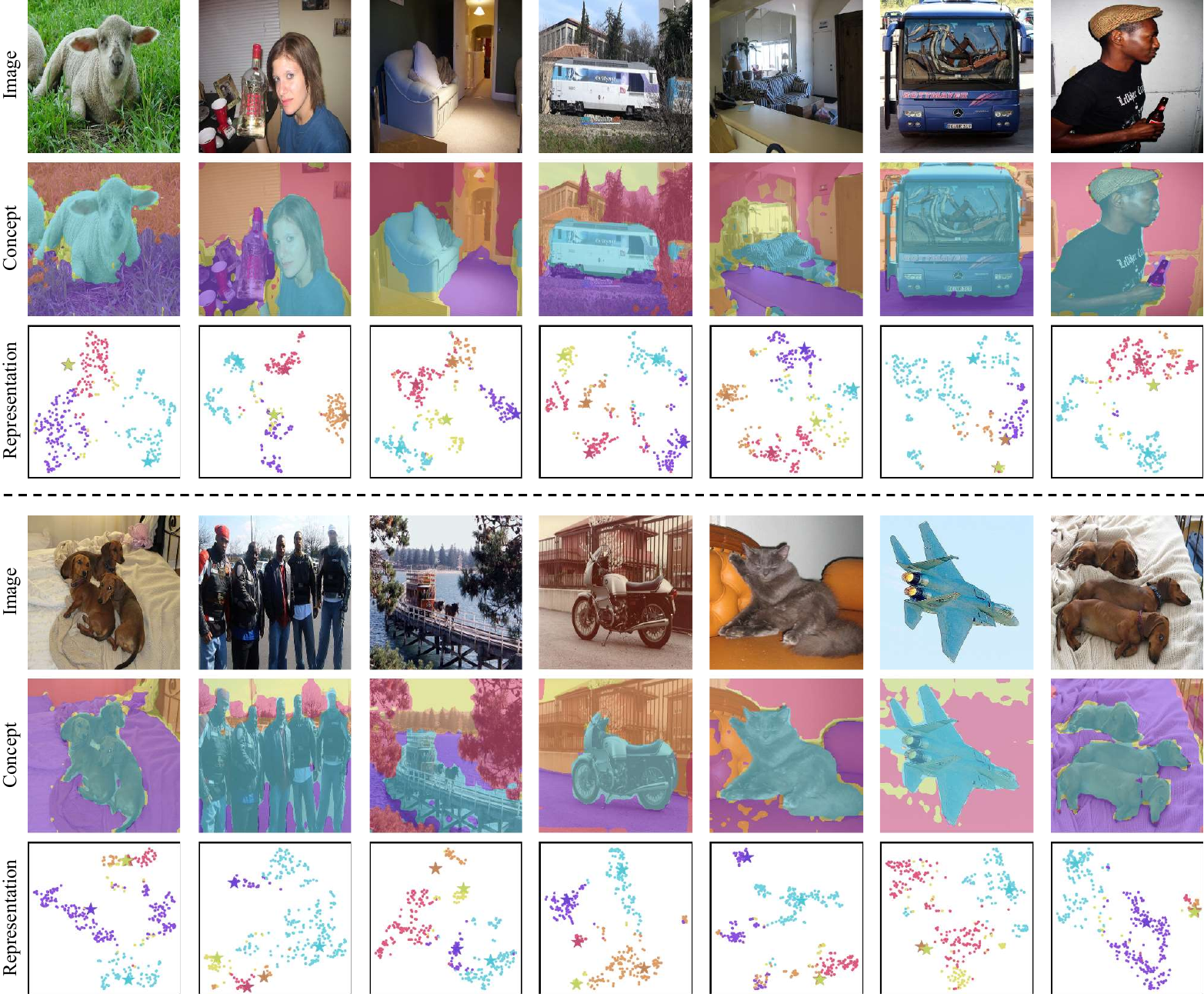}
    \caption{\textbf{Additional t-SNE visualization of the pixel-level representations (marked with dots) produced by self-supervised ViT and the corresponding concepts discovered by the ACG (marked with stars).} We mark the concepts found by the ACG in different colors.}
    \label{fig:appendix_tsne}
\end{figure*}

\begin{figure*}[ht]
    \includegraphics[width=\textwidth]{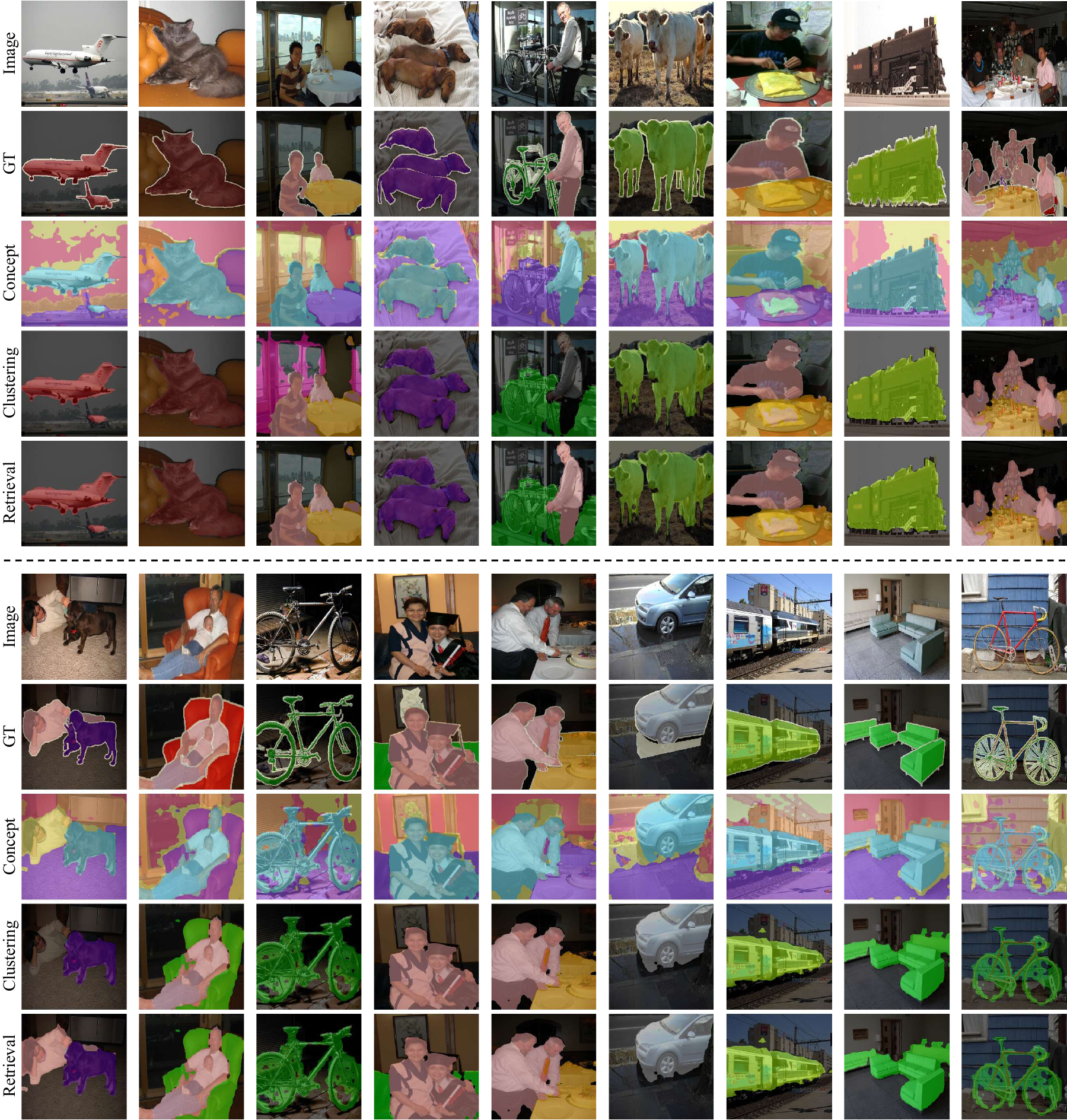}
    \caption{\textbf{Additional qualitative results on PASCAL VOC 2012 dataset.}}
    \label{fig:appendix_clustering}
\end{figure*}

\begin{figure*}[ht]
    \includegraphics[width=\textwidth]{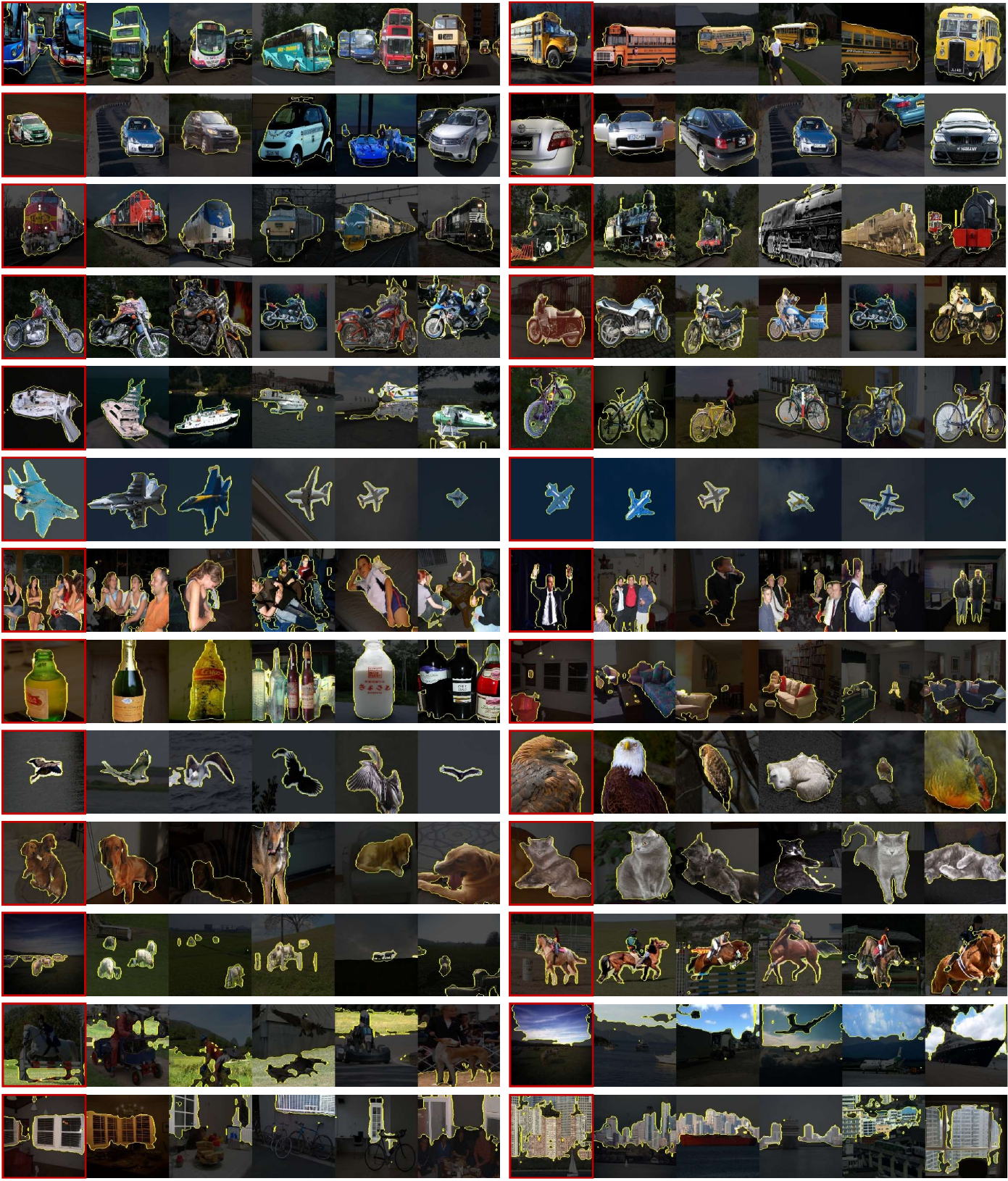}
    \caption{\textbf{Additional visualization of $ k $-NN retrieval results.} We show five concepts with the highest similarity following each query concept (with red frame). The concepts is shown by the highlighted area in the image.}
    \label{fig:appendix_retrieval}
\end{figure*}

We show the additional t-SNE~\cite{van2008visualizing} visualization of the pixel-level representations and discovered concepts, clustering results, and retrieval results in Figure~\ref{fig:appendix_tsne}, Figure~\ref{fig:appendix_clustering}, and Figure~\ref{fig:appendix_retrieval}, respectively.
In addition, the visualization of the training process can be found in \textcolor{blue}{acseg\_video.mp4} in the supplementary material.

\end{document}

%% file: math_commands.tex
\usepackage{amsmath,amsfonts,bm}

\def\eqref#1{equation~\ref{#1}}

\def\1{\bm{1}}

\def\vc{{\bm{c}}}

\def\vx{{\bm{x}}}

\def\mC{{\bm{C}}}

\def\mS{{\bm{S}}}

\def\mW{{\bm{W}}}
\def\mX{{\bm{X}}}

\DeclareMathAlphabet{\mathsfit}{\encodingdefault}{\sfdefault}{m}{sl}
\SetMathAlphabet{\mathsfit}{bold}{\encodingdefault}{\sfdefault}{bx}{n}

\def\sR{{\mathbb{R}}}

\def\emA{{A}}

\def\emS{{S}}

